\documentclass{llncs}
\usepackage{llncsdoc}

\usepackage[utf8]{inputenc}
\usepackage{graphicx}
\usepackage{verbatim}
\usepackage{array}
\usepackage{multirow}
\usepackage{pifont}
\usepackage{amssymb}
\usepackage{mathtext}
\usepackage{fontenc}
\usepackage{xcolor}
\usepackage{colortbl}
\usepackage{textcomp}
\usepackage{url}
\usepackage[linesnumbered,lined,boxed,commentsnumbered,ruled,vlined]{algorithm2e}

\definecolor{ClrCor}{RGB}{20, 48, 255}
\begin{document}

\title{Anveshak - A Groundtruth Generation Tool for Foreground Regions of Document Images}

\author{Soumyadeep Dey \and Jayanta Mukherjee \and Shamik Sural \and Amit Vijay Nandedkar}

\institute{Department of Computer Science and Engineering, Indian Institute of Technology Kharagpur, Kharagpur 721302}

\maketitle

\begin{abstract}
We propose a graphical user interface based groundtruth generation tool in this paper. 
Here, annotation of an input document image is done based on the foreground pixels. 
Foreground pixels are grouped together with user interaction to form labeling units. 
These units are then labeled by the user with the user defined labels. 
The output produced by the tool is an image with an $XML$ file containing its metadata information. 
This annotated data can be further used in different applications of document image analysis.
\end{abstract}


\section{Introduction}
\label{sec:intro}

Document digitization has attracted attention for several years. 
Conversion of a document image into electronic format requires several types of document image analysis. 
Typical document image analysis includes different types of segmentation, optical character recognition ($OCR$), etc. 
Numerous algorithms have been proposed to achieve these objectives. 
The performance of these algorithms can be measured with the help of groundtruth. 
The data with groundtruth is of immense importance in document image analysis. 
It is required for training, machine learning based algorithms, 
and it is also used for evaluation of various algorithms. 
The generation of groundtruth is a manual and time consuming process. 
Hence, the groundtruth generation tool should be user friendly, reliable, effective, 
and capable of generating data in a convenient manner. 

Several systems for groundtruth generation have been reported in the literature 
for producing benchmark datasets to evaluate competitive algorithms. 
Pink Panther~\cite{pinkpanther} is one such groundtruth generator, and is mainly used for evaluation of layout analysis. 
PerfectDoc~\cite{PerfectDoc} is a groundtruth generation system for document images, based on layout structures. 
Various layout based groundtruth generation tools are present in the literature~\cite{trueviz}, \cite{GT_ICDAR2009}, \cite{Li_Das06}. 
These groundtruth generators~\cite{Gford_gt}, \cite{trueviz}, \cite{PerfectDoc}, only support rectangular regions for annotation. 
Hence, they fail to generate groundtruth for documents with complex layout. 

A recent groundtruth generator $GEDI$~\cite{GEDI}, supports annotation by generating a polygonal region. 
However, it is observed that the tool is quite inefficient for images of larger size ($600 dpi$). 
PixLabeler~\cite{PixLabeler} is an example of pixel level groundtruth generator. 
Similar tools are also reported in~\cite{4669984}, \cite{1699028}, \cite{Dori_MVA}. 
Pixel level annotation gives more general measure for annotation, but it involves more time for completing the annotation task. 

In this paper, we propose a tool to annotate a document image at pixel level. 
The main objective of the tool is to efficiently annotate data using less amount of time. 
Towards this, we have provided a semi-automatic interactive platform to annotate document images efficiently. 
Since our main goal is to annotate foreground pixels, we segment foreground pixels from its background with user assistance. 
Next, we group foreground pixels such that neighboring pixels of similar types get connected. 
Finally, annotation of each such group of pixels is performed with a predefined set of labels. 

The system is called Anveshak and its functionality is described in Section~\ref{sec:Functions}. 
Implementation details are discussed in Section~\ref{sec:imple}. 
Section~\ref{sec:GTGen} provides the details of groundtruth generation with Anveshak. 
Finally, we conclude in Section~\ref{sec:conclu}.

\section{Functionality}
\label{sec:Functions}

\begin{figure}[h]
 \centering
 \fbox{\includegraphics[width = .95\textwidth]{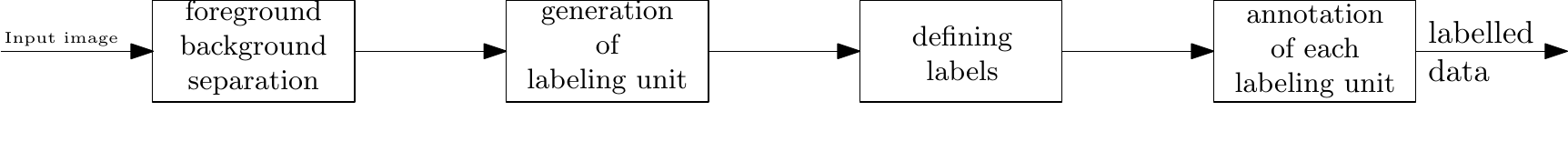}}
 \caption{Work flow of Anveshak system}
 \label{fig:workflow}
\end{figure}

The work-flow of the Anveshak system is shown in Figure~\ref{fig:workflow}. 
Some semi-automated modules are implemented to speed up the annotation process.

\subsection{Foreground Background Separation}
\label{sub-sec:FBS}

\begin{figure}[h!]
 \centering
 \begin{tabular}{@{}c@{\ }c@{}}
\fbox{\includegraphics[width=.9\textwidth]{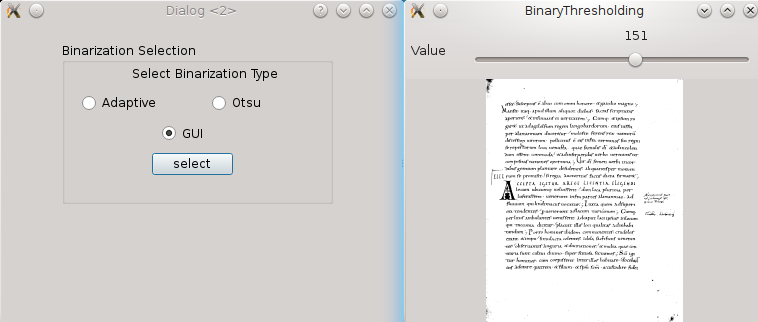}}&
\end{tabular}
\begin{tabular}{@{}c@{\ }c@{}}
\fbox{\includegraphics[width=.9\textwidth]{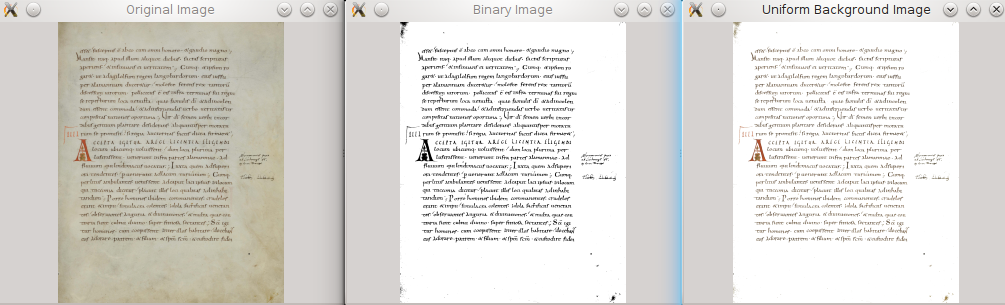}}&
\end{tabular}
\caption{Foreground background separation module.}
\label{fig:binarizationmodule}
\end{figure}

We are mainly concerned with the annotation of foreground pixels of a document image. 
A module is integrated with Anveshak to efficiently segment foreground pixels from its background. 
This task can be performed with three types of thresholding techniques, 
first, $GUI$ based thresholding, second, a $GUI$ based adaptive thresholding technique~\cite{opencv}, 
and third, the \emph{Otsu's} thresholding technique~\cite{otsu}. 
Here, a user can segment foreground from its background efficiently, using either of these three thresholding techniques. 
An example of foreground background separation module using $GUI$ based thresholding is shown in Figure~\ref{fig:binarizationmodule}. 

\subsection{Generation of Labeling Units}
\label{sub-sec:GLU}

Anveshak has a unique technique to predefine labeling units. 
Labeling units are generated using $GUI$ based morphological operations. 
Morphological operations included in Anveshak are, \emph{erosion}, \emph{dilation}, 
\emph{closing}, \emph{opening}, \emph{gap-filling}, and \emph{smoothing}. 

\begin{figure}[h!]
 \centering
 \fbox{\includegraphics[width=.6\textwidth]{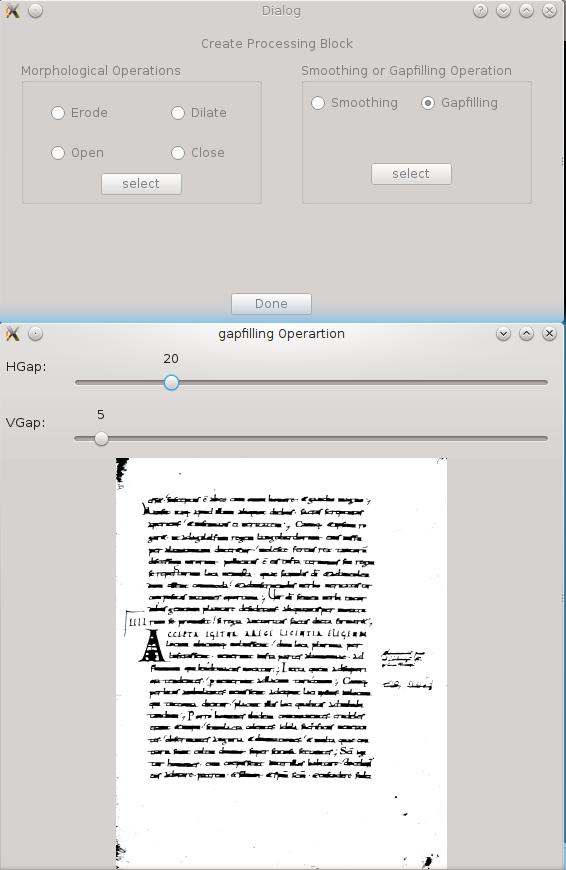}}
 \caption{Module of Anveshak for generating labeling unit with gap filling operation}
 \label{fig:GLU}
\end{figure}

A labeling unit is a collection of foreground pixels, grouped together using a suitable morphological operator. 
Pixels are grouped together by choosing either of these morphological operations - \emph{erosion}, \emph{dilation}, 
\emph{closing}, and \emph{opening}~\cite{gonzalez09}. 
The user can select an ideal element size and element type, in order to group pixels. 
A user can also accumulate pixels to form a group by a smoothing operation~\cite{Wahl1982}, 
where choosing of run length parameter is an interactive process. 
Foreground pixels can also be grouped together using gap filling operation~\cite{sdey12}, 
where selection of the parameter, gap size in horizontal and vertical directions, is a user driven process. 
An instance of Anveshak for generating labeling units is shown in Figure~\ref{fig:GLU}. 

\begin{figure}[h!]
 \centering
 \fbox{\includegraphics[width=.5\textwidth]{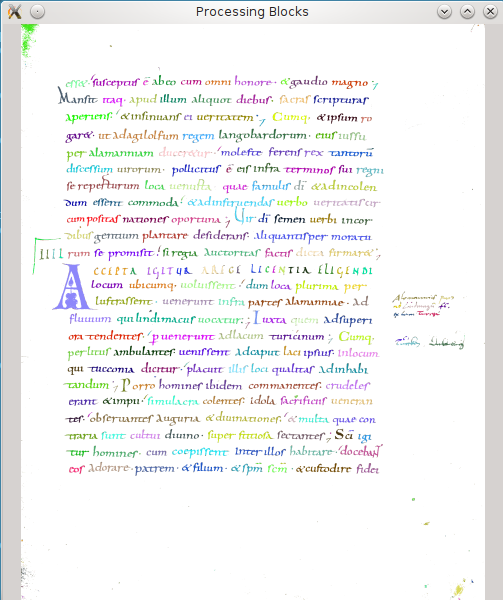}}
 \caption{Labeling units generated by Anveshak with user intervention}
 \label{fig:labelingunit}
\end{figure}

After grouping the pixels, contours of each group is obtained using the method described in~\cite{Suzuki85}. 
Each contour is then approximated to a polygon by applying Douglas-Peucker algorithm~\cite{douglas73}. 
The polygons thus computed are the basic units for annotation in Anveshak. 
An example of a collection of labeling units is shown in Figure~\ref{fig:labelingunit}, 
where each labeling unit is represented using a unique colors. 

\subsection{Defining Labels}
\label{sub-sec:labels}

\begin{figure}[h!]
 \centering
 \fbox{\includegraphics[width=.6\textwidth]{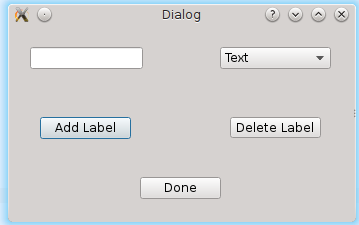}}
 \caption{An instance of Anveshak to set labels}
 \label{fig:setlabels}
\end{figure}

There are some predefined labels in Anveshak. 
The tool provides an option to add and delete labels, as shown in Figure~\ref{fig:setlabels}. 
After defining all the labels, a user can annotate the labeling units of the input document with the defined labels. 
A unique index number and a color is assigned to each label, which are used in the later stages of annotation. 

\subsection{Annotation of Labeling Units}
\label{sub-sec:alu}

\begin{figure}[h!]
 \centering
 \fbox{\includegraphics[width=.9\textwidth]{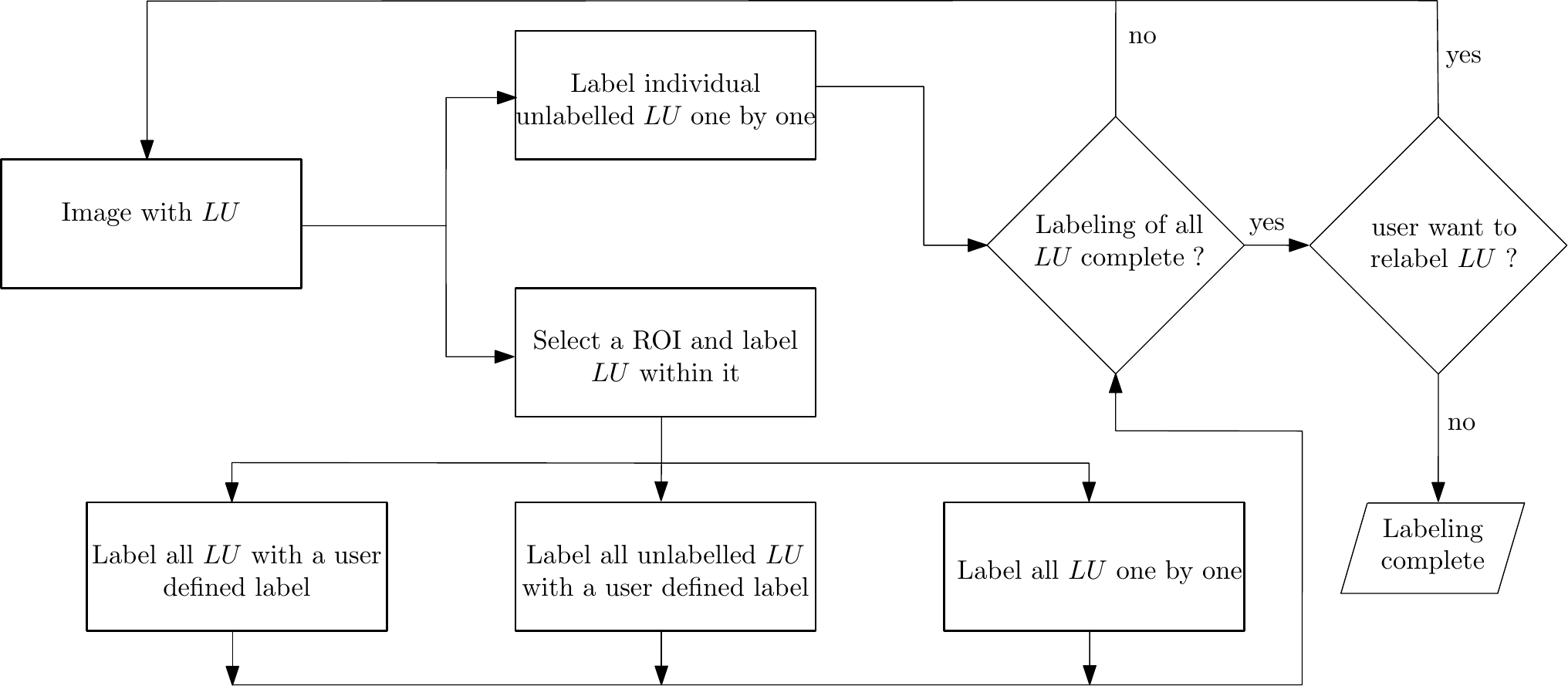}}
 \caption{Work flow of the annotation unit of Anveshak}
 \label{fig:workflow_annotation}
\end{figure}

\begin{figure}[h!]
 \centering
 \fbox{\includegraphics[width=.6\textwidth]{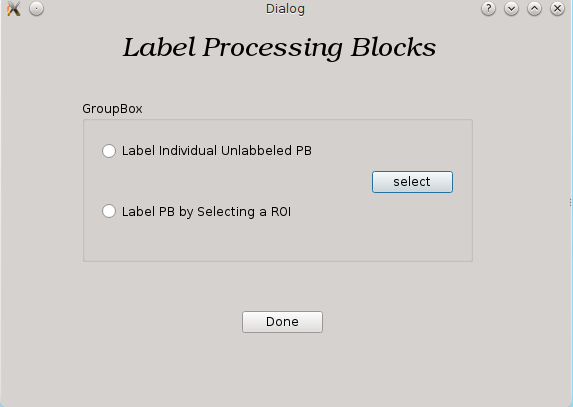}}
 \caption{Instance of Anveshak for choosing annotation mode}
 \label{fig:annotate}
\end{figure}

Overall annotation process can be summarized uasing a flow chart given in Figure~\ref{fig:workflow_annotation}.  
Annotation of labeling units is performed in two ways as shown in Figure~\ref{fig:annotate}. 
A user can label unlabeled units one by one with the predefined labels. 
In this case, an unlabeled unit is displayed in a window and the user is prompted for a label for the displayed unit. 
This process continues until each of the units is labeled, or the user chooses to label the units by selecting a region of interest ($ROI$). 

\begin{figure}[h!]
 \centering
 \fbox{\includegraphics[width=.6\textwidth]{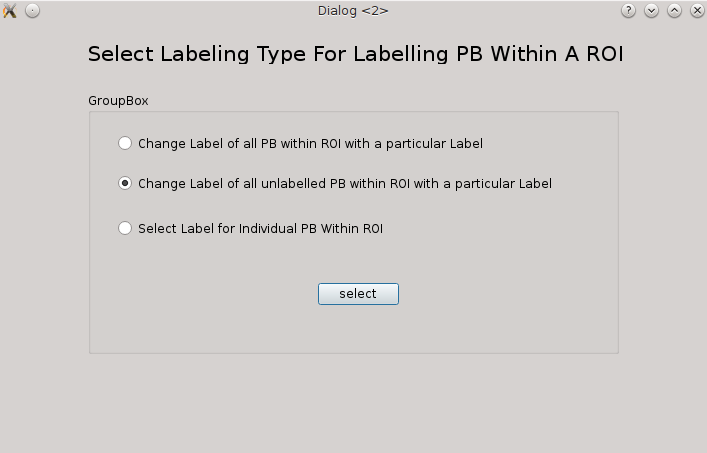}}
 \caption{Options prompted for choosing labeling mode for a $ROI$.}
 \label{fig:ROIoption}
\end{figure}

Another method of labeling units is to select a region of interest. 
In this module, a user can select an $ROI$, which can be annotated with the defined labels. 
At first, all units are determined which are completely present within the selected $ROI$. 
After selection of an $ROI$, units present within the $ROI$ can be labeled using three different modes (Figure~\ref{fig:ROIoption}). 
A user can annotate all units within the $ROI$ with one label, and update all units with the selected label. 
Another way of annotation is by labeling all units belonging to the selected $ROI$ with a particular type. 
Lastly, a user can annotate each unit belonging to the selected $ROI$ individually with a label. 
Pixels belonging to a particular labeling unit are updated with the unique index corresponding to the label of $ROI$, 
and color of those pixels is updated with the color of that label. 
Belongingness of a pixel to a particular labeling unit is computed through point-polygon test. 
At each stage of the annotation process, the updated color image is displayed, 
where labeled pixels are displayed with color of the corresponding label, 
and unlabeled pixels are displayed with their original color value.

\begin{figure}[ht]
 \centering
 \begin{tabular}{@{}c@{\ }c@{\ }c@{}}
\fbox{\includegraphics[width=.3\textwidth]{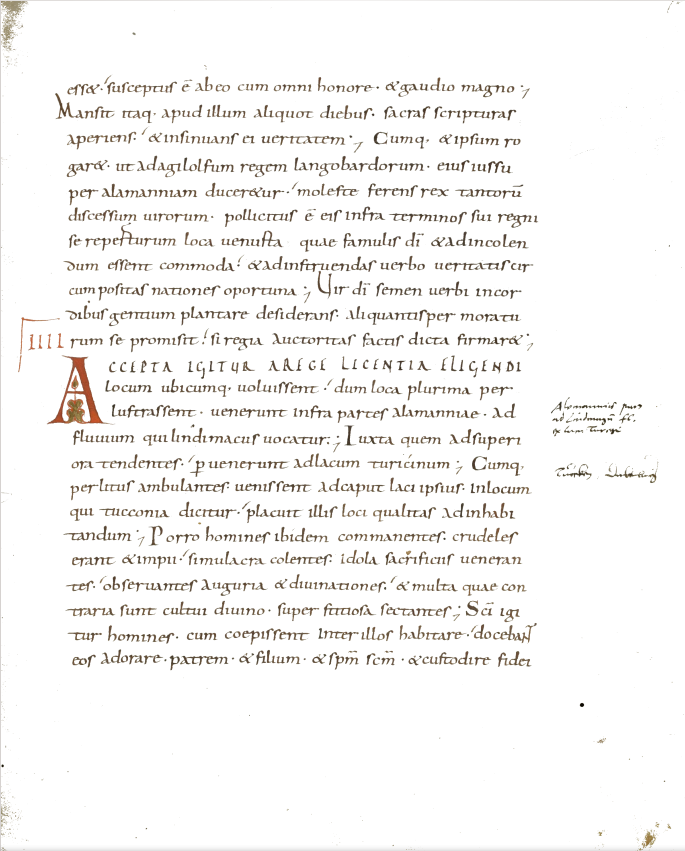}} & 
\fbox{\includegraphics[width=.3\textwidth]{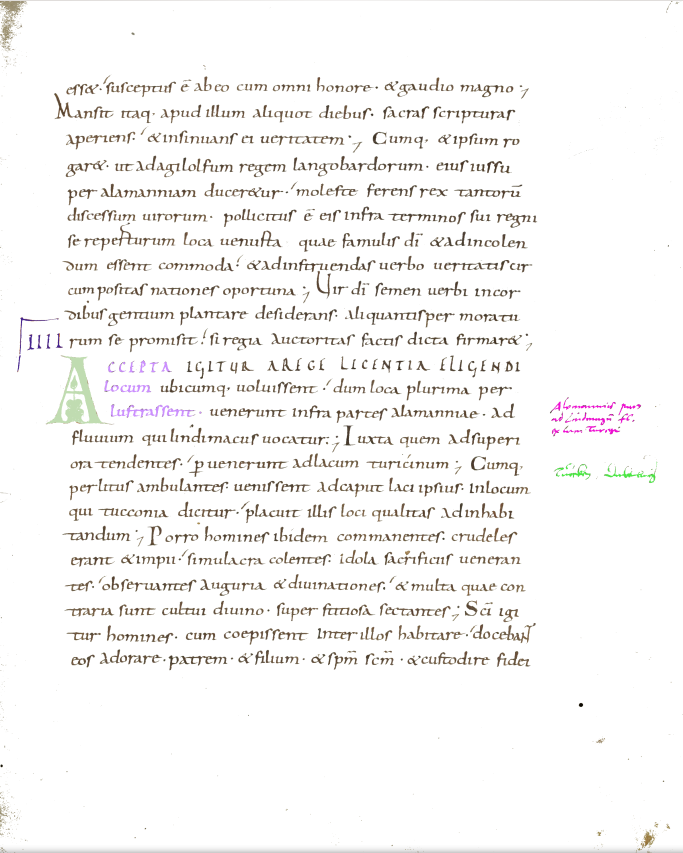}} &
\fbox{\includegraphics[width=.3\textwidth]{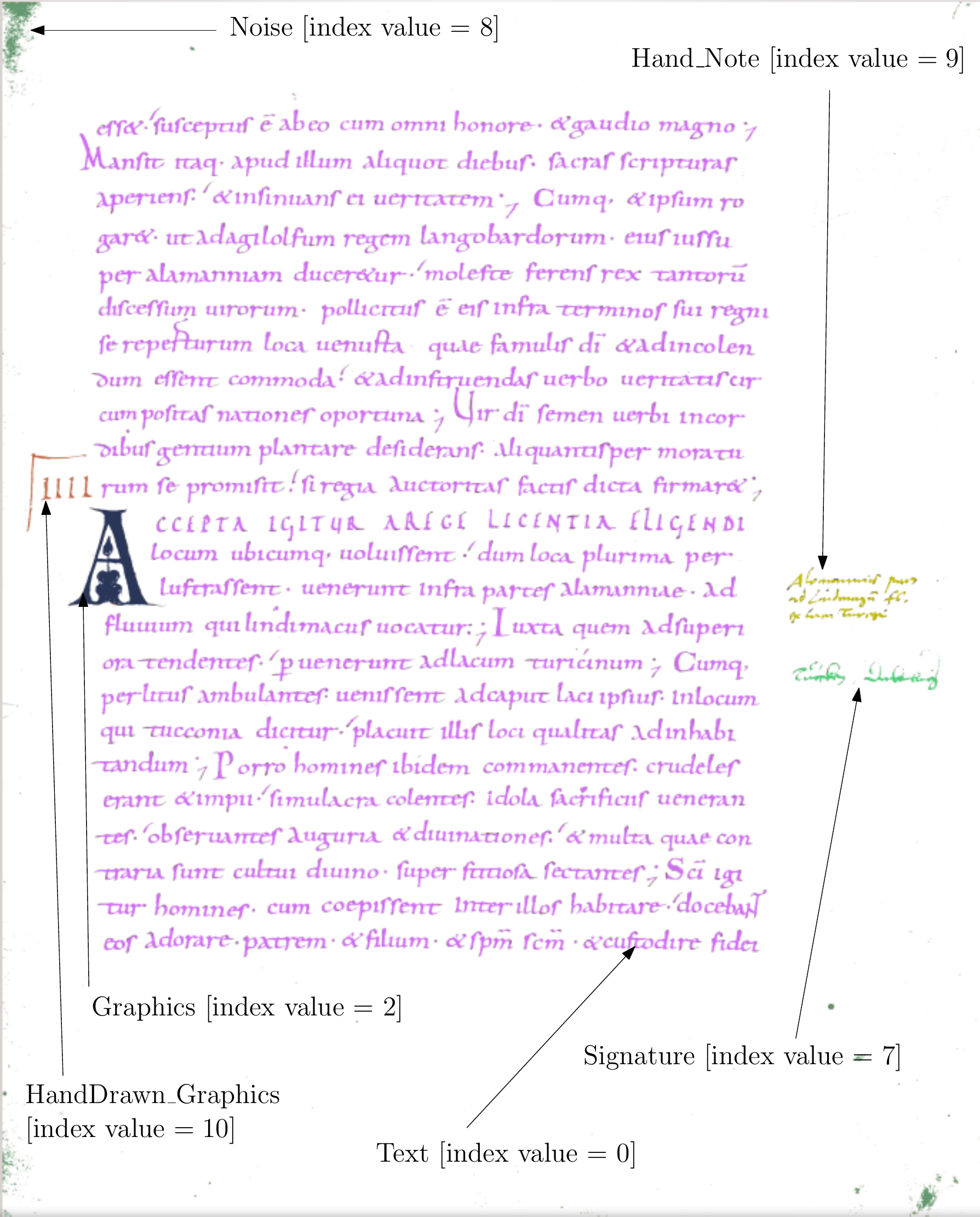}}\\
(a) & (b) & (c) \\
\end{tabular}
\begin{tabular}{@{}c@{\ }c@{}}
\fbox{\includegraphics[width=.75\textwidth]{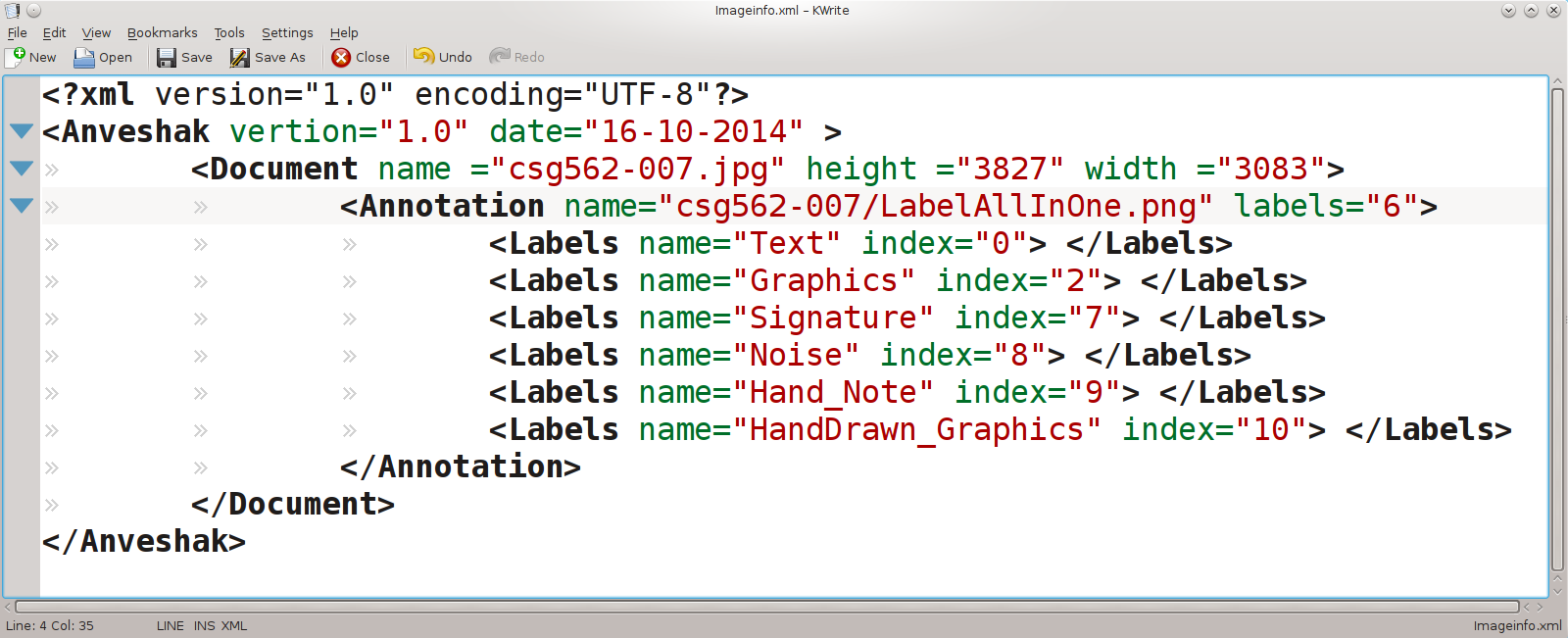}}\\
(d)
\end{tabular}
\caption{(a)~Unlabeled foreground pixels; (b)~Half annotated foreground pixels; 
(c)~Full annotated foreground pixels with a unique color per label; 
(d)~ $XML$ output after annotation of Figure~\ref{fig:labeling}(a)}
\label{fig:labeling}
\end{figure}

The process of annotation continues until all labeling units are marked. 
After completion of annotation, the user is asked, 
whether he/she wants to update any label, or finalize the labels. 
After finalizing the labels, output labeled image and its corresponding $XML$ file are generated. 
An example of different stages of labeling is shown in Figure~\ref{fig:labeling}. 

\section{Implementation Details}
\label{sec:imple}

Anveshak is implemented in $C++$, using cross-platform application framework $Qt$ for graphical user interface 
and with customized modules developed using OpenCV~\cite{opencv}. 
Annotation of an image is achieved through the user interface 
and after completion, a single image in $.png$ format is generated. 
Each pixel of the output image is represented with an index corresponding to a particular annotation. 

\begin{figure}[h!]
 \centering
 \begin{tabular}{@{}c@{\ }c@{}}
\fbox{\includegraphics[width=.45\textwidth]{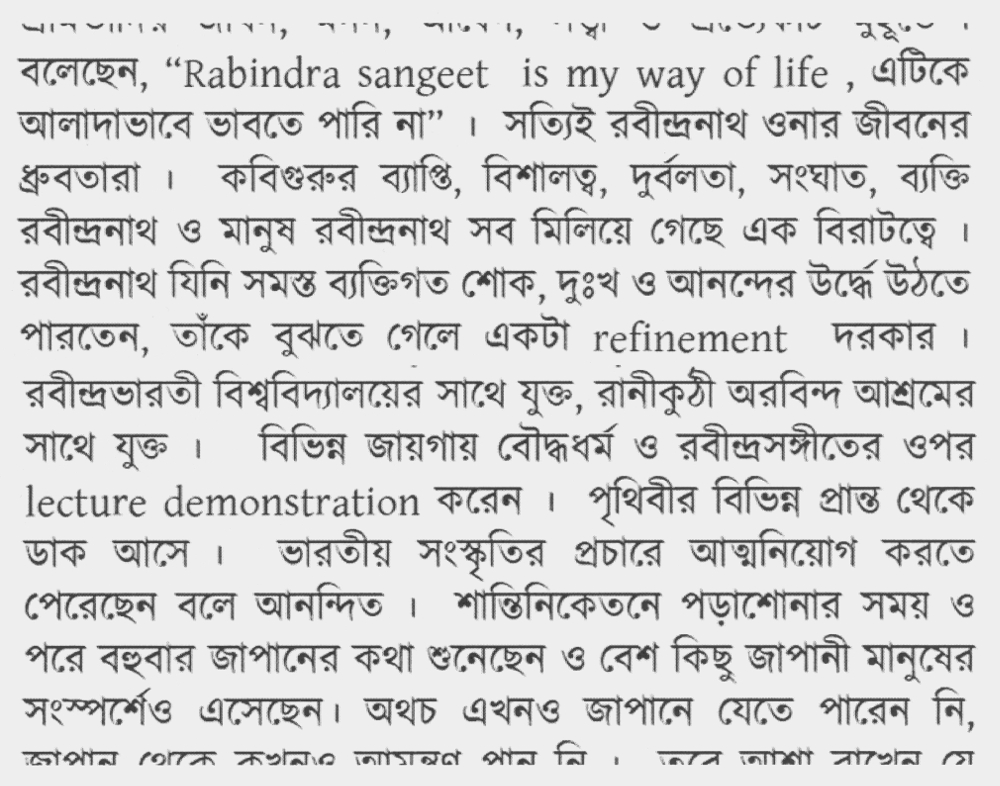}}\\
(a)
\end{tabular}
 \begin{tabular}{@{}c@{\ }c@{}}
\fbox{\includegraphics[width=.45\textwidth]{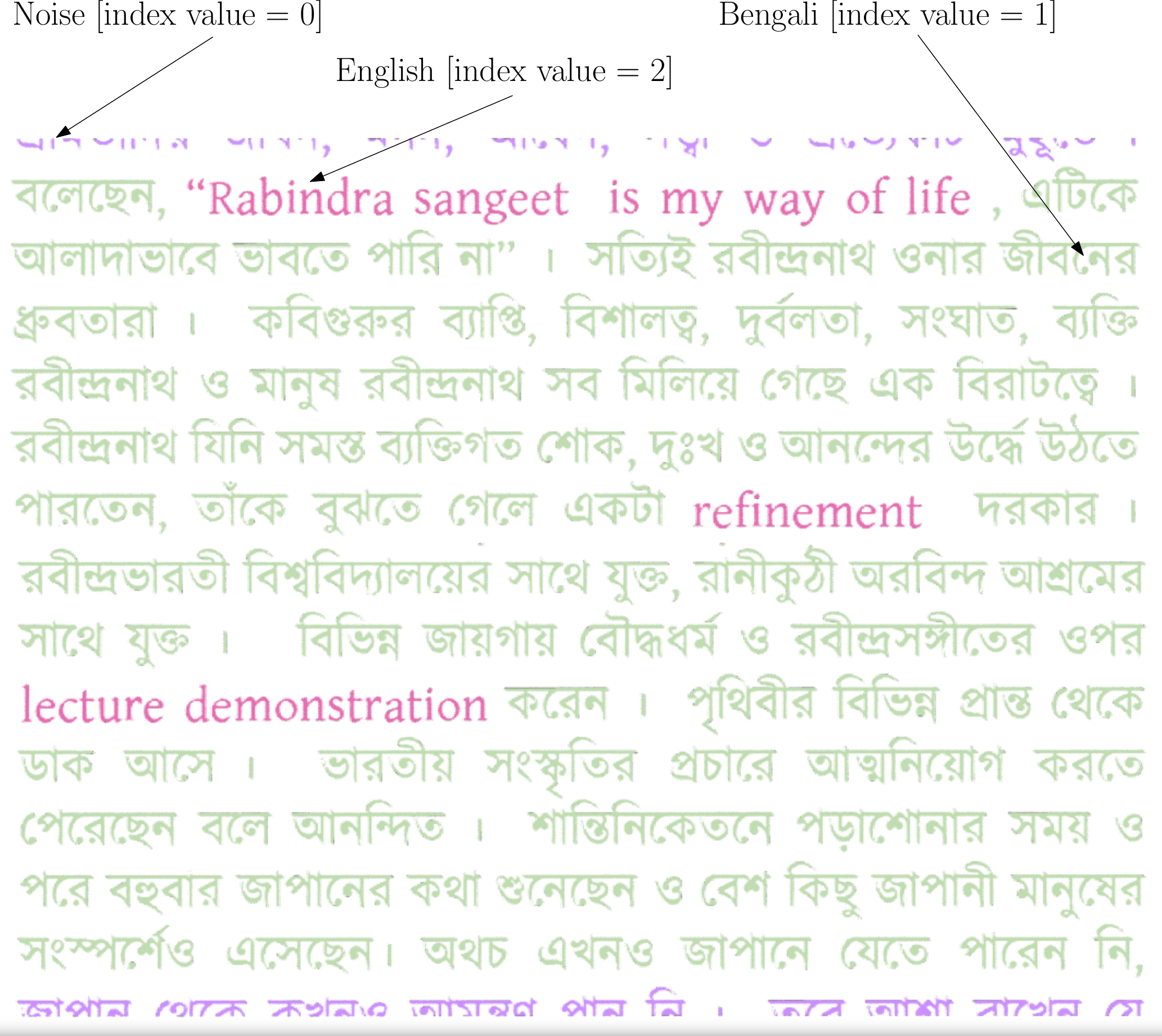}}\\
(b)
\end{tabular}
\begin{tabular}{@{}c@{\ }c@{}}
\fbox{\includegraphics[width=.75\textwidth]{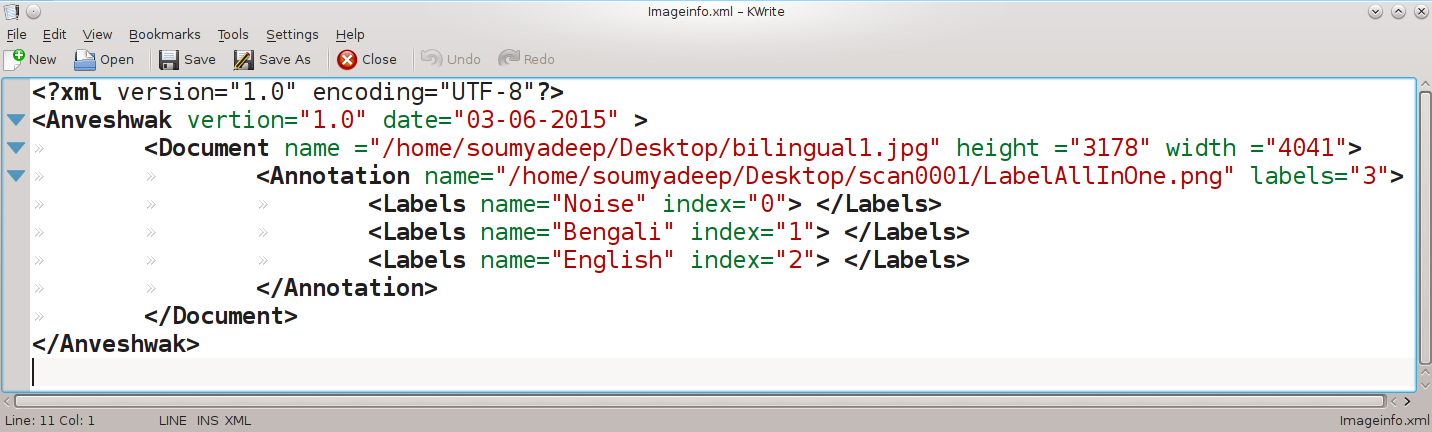}}\\
(b)
\end{tabular}
\caption{(a)~An input bilingual image (b)~Annotated image for Figure~\ref{fig:XMLDATA}~(a) with a unique color per label; 
(b)~$XML$ output after annotation of Figure~\ref{fig:XMLDATA}~(a)}
\label{fig:XMLDATA}
\end{figure}

The metadata of the concerned image is stored in an $XML$ file, 
which also includes the information of the source image along with the annotated image. 
In the $XML$ file, an index corresponds to the unique pixel value for a particular label in the annotated image. 
Examples of two different annotated images and their corresponding $XML$ files 
are respectively shown in Figures~\ref{fig:labeling}~(c) and (d) and Figures~\ref{fig:XMLDATA}~(a) and~(b). 
Anveshak is tested to annotate $344$ images from the dataset reported in~\cite{Micenkova:11ICDAR:stamp}.  
It has been observed by the annotator that, the labeling can be performed in a much easier and faster way than 
it could be performed with PixLabeler~\cite{PixLabeler} or $GEDI$~\cite{GEDI}.

In our present implementation of Anveshak, only one annotation per block is supported. 
In many scenarios, it is desirable to have multiple annotations per block, mainly in case of overlapping regions. 
In future, we plan to support more than one annotation per block. 
Present implementation of Anveshak has been made available online\footnote{\url{http://www.facweb.iitkgp.ernet.in/~jay/anveshak/anveshak.html}}. 


\section{Generation of Groundtruth using Anveshak}
\label{sec:GTGen}

Anveshak is used to generate groundtruth for the dataset reported in~\cite{Micenkova:11ICDAR:stamp}. 
The images in the dataset consist of various regions like logo, headers, text, signature, headline, bold text, etc. 
However, annotation of stamp regions is only available with the original dataset. 
The dataset consists of $425$ scanned images in $600$, $300$, and $200$ $dpi$ resolutions. 
Out of these $425$ images, $344$ images contain non overlapping regions. 
Anveshak is used to annotate these $344$ images of $300$ $dpi$ resolution, 
and the groundtruth data has been made available online\footnote{\url{http://www.facweb.iitkgp.ernet.in/~jay/anveshak_gt/anveshak_gt.html}}. 
These $344$ images are annotated using Anveshak with the help of $6$ users. 
There are on an average $5$ labels, and $148$ segments per image in the given dataset. 
Users involved in annotation are initially trained to annotate data with one random image. 
Average time taken by a user to annotate an image with Anveshak is about $3-4$ minutes. 
The annotated dataset has been used in the works reported in~\cite{sdey_stamp_NCVPRIPG} and \cite{Dey2016_IJDAR}.

\section{Conclusion}
\label{sec:conclu}

The primary target of Anveshak is to annotate an input document image in an efficient manner.  
Our tool produces an $XML$ file containing the metadata information, along with an annotated image. 
We have developed a user friendly groundtruth generation tool, 
with some semi-automatic modules which make the annotation process faster. 
We hope that Anveshak will serve the document analysis community in an effective manner 
by simplifying groundtruth generation procedure.

 \section*{Acknowledgments}
This work is partly funded by TCS research scholar program and partly by Ministry of Communications \& Information Technology, Government of India; 
MCIT 11(19)/ 2010-HCC (TDIL) dt. 28-12-2010.

\end{document}